\documentclass[10pt,twocolumn,letterpaper]{article}

\usepackage{cvpr}
\usepackage{times}
\usepackage{epsfig}
\usepackage{graphicx}
\usepackage{amsmath}
\usepackage{amssymb}

\graphicspath{ {images/} }
\usepackage{subcaption}
\graphicspath{ {images/} }
\usepackage{float}
\usepackage{placeins}

\newcommand*{\authorimg}[1]{%
  \raisebox{-.1\baselineskip}{%
    \includegraphics[
      height=\baselineskip,
      width=\baselineskip,
      keepaspectratio,
      trim={0.31cm 0 0.31cm 0},
      clip
    ]{#1}%
  }%
}

\usepackage{xcolor,url}
\usepackage{soul}

\usepackage[breaklinks=true,bookmarks=false]{hyperref}

\cvprfinalcopy % *** Uncomment this line for the final submission

\def\cvprPaperID{6701} % *** Enter the CVPR Paper ID here

\begin{document}

%%%%%%%%% TITLE
\title{Deep Variational Transfer: Transfer Learning through Semi-supervised Deep Generative Models}

%, Harvard University, Cambridge, MA, USA

\author{Marouan Belhaj\\
IACS, Harvard University\\
Cambridge, MA, USA\\
{\tt\small blhmarouan@gmail.com}
\and
Pavlos Protopapas\\
IACS, Harvard University\\
Cambridge, MA, USA\\
{\tt\small pavlos@seas.harvard.edu}
\and
Weiwei Pan\\
IACS, Harvard University\\
Cambridge, MA, USA\\
{\tt\small weiweipan@g.harvard.edu}
}

\maketitle
%\thispagestyle{empty}

%%%%%%%%% ABSTRACT
\begin{abstract}
    In real-world applications, it is often expensive and time-consuming to obtain labeled examples. In such cases, knowledge transfer from related domains, where labels are abundant, could greatly  reduce the need for extensive labeling efforts. In this scenario, transfer learning comes in hand. 
    In this paper, we propose Deep Variational Transfer (DVT), a variational autoencoder that transfers knowledge across domains using a shared latent Gaussian mixture model. Thanks to the combination of a semi-supervised ELBO and parameters sharing across domains, we are able to simultaneously: (i) align all supervised examples of the same class into the same latent Gaussian Mixture component, independently from their domain; (ii) predict the class of unsupervised examples from different domains and use them to better model the occurring shifts.
    We perform tests on MNIST and USPS digits datasets, showing DVT's ability to perform transfer learning across heterogeneous datasets. Additionally, we present DVT's top classification performances on the MNIST semi-supervised learning challenge. We further validate DVT on a astronomical datasets. DVT achieves states-of-the-art classification performances, transferring knowledge across real stars surveys datasets, EROS, MACHO and HiTS, . In the worst performance, we double the achieved F1-score for rare classes.
    These experiments show DVT's ability to tackle all major challenges posed by transfer learning: different covariate distributions, different and highly imbalanced class distributions and different feature spaces.
\end{abstract}

%%%%%%%%% BODY TEXT

\section{INTRODUCTION}

In supervised learning, we learn a prediction function $f: \mathcal{X} \rightarrow \mathcal{Y}$ from a training set $D_{S} = \left \{ (x_{1}^{S},y_{1}^{S}),...,(x_{n}^{S},y_{n}^{S}) \right \} \subseteq \mathcal{X} \times \mathcal{Y}$, where $\mathcal{X}$ and $\mathcal{Y}$ respectively denote the space of predictors $X$ and labels $Y$. The learned $f$ is expected to generalize well on a test set under the assumption that training and test sets are drawn from the same distribution. When this assumption fails, the data is said to have undergone \emph{dataset shift}, i.e., $p_{S}(x,y) \neq p_{T}(x,y)$. In these cases, the test predictions of $f$ cannot be trusted. One solution might be to acquire new labels for data similar to the test set and retrain the model. However, in many cases, acquiring new labels as well retraining might be prohibitively expensive with respect to time and cost. In \emph{transfer learning}, our goal is to build models that adapt and generalize well even when data distributions between train and test sets differ. 

In this paper, we develop a method to address transfer learning under \emph{general dataset shifts}, defined as a change in both factors in the joint distribution $p(x,y)=p(x|y)p(y)$ (respectively $p(x,y) = p(y|x)p(x)$) as one moves from training to test data \cite{moreno2012unifying}. In astronomy, for example, models trained on celestial objects labeled from one night sky survey often cannot be directly applied to classify objects in new surveys, since each survey presents variable conditions such as different  atmospheric conditions, sensor sensitivity etc. 

Methods in existing literature on transfer learning often assume unsupervised transfer learning conditions, where no target label is available. Under this hypothesis, it is extremely hard if not impossible to solve for general dataset shifts. This drives existing literature towards two possible choices: (i) restricting the types of shifts taking place and solving for one of three simpler shifts: covariate, prior or conditional shift; (ii) consider general shifts while restricting the type of transformations allowed, thus solving for location-scale conditional shifts only. Very few work concentrates on semi-supervised transfer learning conditions, with lack of methods able to simultaneously deal with high-dimensional data and heterogeneous domains $\mathcal{X}_{S} \neq \mathcal{X}_{T}$.

In this paper, we propose a novel semi-supervised transfer learning model, Deep Variational Transfer (DVT), for classification under general dataset shifts. Based on \emph{source} labels and few \emph{target} labels, our model exploits parameters sharing to learn a latent projection of the data wherein each class, independent of domain, is aligned with a unique and pre-determined component of a latent Gaussian mixture model. This allows us to effectively correct for changes in class-conditional distributions. 

We exploit supervised samples to train domain specific classifiers $q_{\phi}(y|\mathbf{z})$ directly from latent projections $z$ of the covariates $x$, allowing DVT to correct for differences class distributions $p_{S}(y) \neq p_{T}(y)$. The classifiers are simultaneously trained on supervised data and used to classify unsupervised data points. In this way, we can decide towards which latent Gaussian mixture component we want to drive our unsupervised points. 
%\pp{The latent variable z is not introduced. Actually you should mention that the classifier is in the latent space. Something is missing at the begining of this paragraph}

This idea works effectively thanks to a reconstruction term that forces the representation of supervised and unsupervised samples close in input space to be close in latent space too. While we push supervised samples toward their pre-determined latent Gaussian mixture component, the unsupervised samples close in latent space follow along, ultimately getting the unsupervised samples to be correctly classified.

\section{Related Work}

%[\textbf{Introduce your notations here (be BRIEF!), for example: We suppose that we have two sets of data $D_S$ and $D_T$ from a source and target distribution $p_S(x, y)$ and $p_T(x, y)$, respectively. Let $p_{S}(y|x)$ be the likelihood on the source data ..}]

Suppose we have a source set $D_{S} = \left \{ (x_{1}^{S},y_{1}^{S}),...,(x_{n}^{S},y_{n}^{S}) \right \} \subseteq \mathcal{X} \times \mathcal{Y}$ and a target set $D_{T} = \left \{ (x_{1}^{Tts},y_{1}^{T}),...,(x_{m}^{T},y_{m}^{T}) \right \} \subseteq \mathcal{X} \times \mathcal{Y}$, where $\mathcal{X}$ and $\mathcal{Y}$ respectively denote the domains of predictors $X$ and classes $Y$. The two sets are respectively drawn from a source and target distributions $p_S(x, y)$ and $p_T(x, y)$. Dataset shift can be described by some combination of three phenomena. \textit{Covariate shift} \cite{shimodaira2000improving, sugiyama2008direct, huang2007correcting} is defined as a change in the joint distributions that is solely attributable to a change in the marginal distribution of the covariates, $p_{S}(y|x) = p_{T}(y|x)$ but $p_{S}(x) \neq p_{T}(x)$. \textit{Prior shift} \cite{storkey2009training} or \textit{class imbalance} \cite{jiang2008literature} arises when the marginal distributions of the classes changes, while the conditional distribution of the features given the classes stay the same, $p_{S}(x|y) = p_{T}(x|y)$ but $p_{S}(y) \neq p_{T}(y)$. \textit{Conditional shift} \cite{moreno2012unifying} or \textit{concept drift} \cite{widmer1996learning} takes place when $p_{S}(x) = p_{T}(x)$ but $p_{S}(y|x) \neq p_{T}(y|x)$ ($X \rightarrow Y$) or $p_{S}(y) = p_{T}(y)$ but $p_{S}(x|y) \neq p_{T}(x|y)$ ($Y \rightarrow X$). 

\paragraph{Unsupervised Transfer}
The vast majority of existing transfer learning methods work under unsupervised conditions where target labels  $y_{i}^{T}$ are unknown. To solve such a complex task, restrictions on the data shifts are imposed, often solving for one of the presented shift conditions at the time. However, handling simpler shifts one at the time is not sufficient in presence of \emph{generic dataset shift}. In fact, many real case scenarios present changes in the join distribution $p(x_{S},y_{S}) \neq p(x_{T},y_{T})$ due to combinations of the presented shifts tacking place at the same time.

In \cite{long2016deep} the authors propose the Joint Maximum Mean Discrepancy (JMMD) as measure of the discrepancy between two join distributions. This measure directly extends the Maximum Mean Discrepancy (MMD) proposed by \cite{gretton2009covariate}. JMMD uses MMD to re-weight the source datapoints  to match the covariate distributions between the source and the target sets in a high dimensional feature space, specifically, a reproducing kernel Hilbert space (RKHS).  However, MMD does not measure the discrepancy in joint distributions $p_{S}(x,y)$ and $p_{T}(x,y)$,  because MMD has not been directly defined for joint distributions in \cite{gretton2009covariate, gretton2012kernel}. \cite{long2016deep} assumes a $X \rightarrow Y$ causal relation and extends MMD to match changes in covariate distribution $p(x)$ and conditional distribution $p(y|x)$. However, in the experiments the authors use datasets presenting equal class distributions $p_{S}(y)=p_{T}(y)$, implicitly assuming no prior shifts. Factoring $p(x,y)$ as $p(y)p(x|y)$, we see that fixing $p(y)$ is equivalent to say that the simultaneous changes in $p(x)$ and $p(y|x)$ are caused by conditional shifts in $p(x|y)$ only. This is a restrictive assumption that does not allow for modeling general domain shifts.

In general, the simultaneous correction of conditional and target shifts under unsupervised regime is an ill-posed problem. Therefore, \cite{zhang2013domain} enforces an additional condition for which the conditional distribution may only change under location-scale transformations on $x$. This makes the problem tractable even in the fully unsupervised case. Yet, this assumption might still be too restrictive for experiments like digits (see section \ref{ssec:digits-experiment}), where the datasets live in two heterogeneous input spaces, and stars (see section \ref{ssec:stars-experiment}) where non-linear transformations across domains might not be correctly modeled assuming location-scale transformations.

\cite{benavente2017automatic} propose a full probabilistic model that represents the joint distribution of features from two domains and simultaneously moels a probabilistic transformation as well. However, the unsupervised conditions forces the authors considering only simple transformations, translation, rotation and scaling, equivalent to the location-scale assumption made in \cite{zhang2013domain}.

\paragraph{Semi-supervised Transfer}

%\begin{figure}
%  \centering
%  \includegraphics[width=\linewidth]{3_mockup_low_density_boundary.png}
%  \caption{When supervised examples are not enough to correctly model and separate class distributions, the supervised decision boundary fails to generalize (violet line). Instead, semi-supervised techniques exploit unlabeled data to better model the data distribution, thus drawing a better decision boundary with the same set of supervised examples (orange line).}
%  \label{fig:mockup_semi_supervised}
%\end{figure}

In many real-world use cases we might work in semi-supervised transfer learning conditions, where few target labels are available. This is often the case, even in situation where collecting new fully supervised datasets is too complex or expensive. In such situations, semi-supervised transfer learning helps to exploit the abundance of unlabeled data, on top of few labeled examples, to improve classification and regression tasks.

%Intuitively, in semi-supervised classification we would like to: (i) keep examples from different classes in different regions and (ii) have classification boundaries dividing low density areas of the unsupervised examples. See figure \ref{fig:mockup_semi_supervised} for an intuitive example.

\cite{wang2014flexible} propose to learn a nonlinear mapping to simultaneously align $p(x)$ and $p(y)$. After assuming a mild smoothness transformation condition, they propose a location-scale transformation parametrized by a non-linear kernel and impose a Gaussian Process prior to make the estimation stable. The non-linearity of the transformation makes their approach able to model a wider set of changes, effectively supporting transformations more complex then in \cite{zhang2013domain} at the cost of few target labels. However, their objective function minimizes the discrepancy between the transformed labels and the predicted labels for only the labeled points in the test domain. While they correct for this limitation by adding a regularization term on the transformation for all $D_{T}$, the two terms are not directly comparable, making the regularization weight hard to set. Furthermore, the method cannot be directly extended to heterogeneous situations, where the space of source and target datasets undergoes changes $\mathcal{X}_{S} \neq \mathcal{X}_{T}$. An example of such situation is presented in section \ref{ssec:digits-experiment}, where the MNIST and USPS digit datasets present a different input dimensionality.

%\textbf{[Summarize semi-supervised transfer HERE VERY BRIEFLY (including the semi-supervised VAE stuff)! Emphasize how your work differs.]}

\section{Background}
%\textbf{Explain how mixture models are used in semi-supervised learning!}

\paragraph{Mixture models} We built DVT by combining ideas from various literature works. From semi-supervised \textit{mixture models}, we would like to reuse one simple idea: assign each class to a mixture component and model data points' likelihood under these components. If we were able to model such a mixture of distributions on the target through few labels, we would be also able to model changes between source and target. Ultimately, we would transfer knowledge from the source domain and achieve performance improvements.

Semi-supervised mixture models \cite{nasrabadi2007pattern} define a family $\left \{p(\mathbf{x}|y, \boldsymbol{\theta}) \right \}$ to represent class distributions $p(\mathbf{x}|y)$. These distributions are then weighted based on a mixture weight vector $\pi_{y}$. A natural criterion modelling the joint log likelihood $\mathcal{L}(\theta; \mathcal{D}^{sup}, \mathcal{D}^{unsup})$ can be derived as:

\begin{equation} \label{eq:mixture_semi_sup}
\setlength\abovedisplayskip{0pt}
    \centering
    \displaystyle
    \begin{aligned} 
      & \sum_{i=1}^{l} \textrm{log} \; \pi_{y_{i}} p(\mathbf{x}_{i}|y_{i},\boldsymbol{\theta}) + \gamma \sum_{i=l+1}^{l+u} \textrm{log} \; \sum_{y=1}^{K} \pi_{y} p(\mathbf{x}_{i}|y_{i},\boldsymbol{\theta}),
    \end{aligned}
\end{equation}

\noindent where $u$ is the number of unsupervised data points $\mathcal{D}^{unsup} = \left \{ x_{1},...,x_{u} \right \}$ and $l$ the number of supervised data points $\mathcal{D}^{sup} = \left \{ (x_{1}, y_{1}),...,(x_{l}, y_{l}) \right \}$ and $\gamma$ allows for weighting the importance of supervised and unsupervised terms. We now transformed the semi-supervised problem into a maximum likelihood (ML) estimate in the presence of missing data, where we treat $y$ as a latent variable that we want to marginalize for in case of unlabeled data.

The first important limitation of the presented mixture model is the curse of dimensionality. For an increasing number of dimensionality, we need an exponentially increasing number of samples in order to reliably estimate the terms in eq. \ref{eq:mixture_semi_sup}. The second limitation comes with the choice of a parametric class of distributions that often boils down to standard and conjugate distributions. While these assumptions simplify inference and learning, they do not represent the real data distribution. 

%To overcome this problem, we need a method which reduces the data dimensionality. Once such reduction is performed, we would use the transformed low dimensional coordinates to reliably perform density estimations and exploit the clustering assumption: examples belonging to the same class should be close in feature space.

%To overcome this problem, we would like to handle flexible and potentially non-standard parametric distributions $p(\mathbf{x}|y)$, allowing for a better fit of each class distribution.

\paragraph{Variational Autoencoder}

Variational autoencoders \cite{kingma2013auto, rezende2014stochastic} offer a possible solution to both avoiding curse of dimensionality and modelling non-standard probability distributions. VAEs, differently from semi-supervised mixture models, works under unsupervised conditions, where the training datasets $\mathcal{D}^{unsup} = \left \{ x_{1},...,x_{u} \right \} \subseteq \mathcal{X}$ consists of $u$ i.i.d. samples of an observed random variable $x$. One key step of VAE is to evaluate the likelihood of a single data point $p_{\theta}(\mathbf{x}_{i})$ as:

\begin{equation}
\setlength\abovedisplayskip{0pt}
    \centering
    \displaystyle
    \begin{aligned} 
        & \textrm{log} \;  p_{\theta}(\mathbf{x}_{i}) \\
      & = \mathcal{L}^{ELBO}(\phi, \theta; \mathbf{x}_{i}) + D_{KL}(q_{\phi}(\mathbf{z}|\mathbf{x}_{i})||p_{\theta}(\mathbf{z|x}_i)).
    \end{aligned}
\end{equation}

where the KL-divergence is intractable. VAE exploits the KL-divergence non-negativity to define a lower bound on the log-likelihood of a single data point, the evidence lower bound (ELBO):

\begin{equation} \label{eq:l_elbo_vae}
\setlength\abovedisplayskip{0pt}
    \centering
    \displaystyle
    \begin{aligned} 
        & \textrm{log} \;  p_{\theta}(\mathbf{x}_{i})  \geq \mathcal{L}^{ELBO}(\phi, \theta; \mathbf{x}_{i}) \\
      & = -D_{KL}(q_{\phi}(\mathbf{z}|\mathbf{x}_{i})||p_{\theta}(\mathbf{z})) + \mathbb{E}_{q_{\phi}(\mathbf{z}|\mathbf{x}_{i})} \left [ \textrm{log} \; p_{\theta}(\mathbf{x}_{i}|\mathbf{z}) \right ],
    \end{aligned}
\end{equation}

where the encoder $q_{\phi}(\mathbf{z}|\mathbf{x})$ and the decoder $p_{\theta}(\mathbf{z}|\mathbf{x})$ are set to be Gaussian distributions parametrized by arbitrarily flexible neural networks, respectively with parameters $\left \{ \phi,\theta \right \}$. $p_{\theta}(\mathbf{z})$ is defined to be a simple Isotropic Gaussian, so as to retrieve an analytical solution for the KL-divergence in eq. \ref{eq:l_elbo_vae}. To optimize over $\mathbb{E}_{q_{\phi}(\mathbf{z}|\mathbf{x}_{i})} [\textrm{log} \; p_{\theta}(\mathbf{x}_{i}|\mathbf{z})]$, VAE defines the reparametrization trick: moving the sampling operation, a non-continuous operation for which gradients cannot be computed, to an input layer. This allows to calculate gradients with respect to both encoder and decoder parameters as the expectation of simple gradients \cite{kingma2013auto, rezende2014stochastic}. At this point, stochastic gradient ascent is performed on the evidence lower bound $\mathcal{L}^{ELBO}$ to jointly train encoder and decoder networks, ultimately maximing the log-likelihood of our data points.

VAE offers two main advantages: (i) it achieves robustness against the curse of dimensionality by projecting similar data points into latent factors $\mathbf{z}$ that are hopefully close in latent space, even when $\mathcal{X}$ is high-dimensional; (ii) it models arbitrarily flexible probability distributions by means of neural networks. Disadvantage is that VAE does not consider information about labelled data.

We would be interested in combining VAE and semi-supervised mixture model qualities in one framework, all while adding transfer capabilities.

\section{Deep Variational Transfer}

\begin{figure*}
  \includegraphics[width=\textwidth,trim={0 0 1cm 0.1cm},clip]{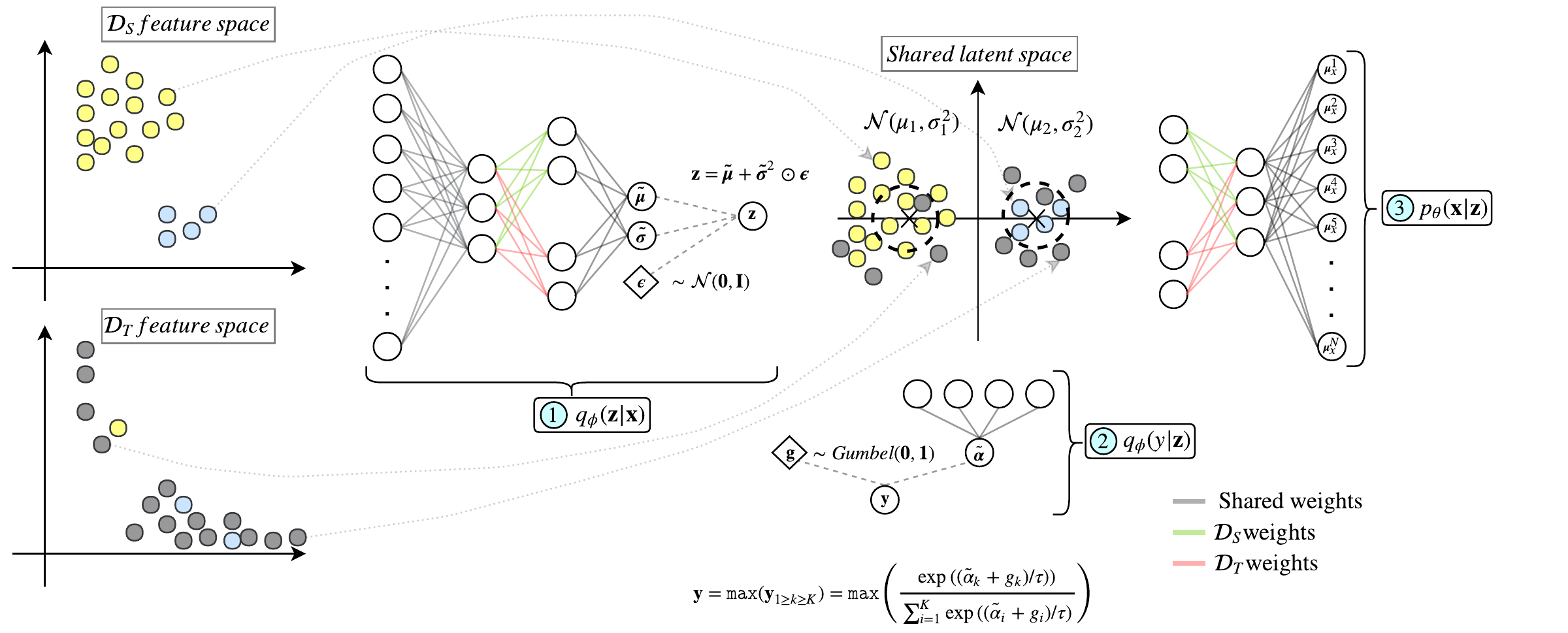}
  \caption{.}
  \label{fig:dvt_overview}
\end{figure*}

We propose Deep Variational Transfer (DVT), a model facilitating transfer of knowledge across domains by aligning them in a shared latent Gaussian mixture model. 

As illustrated in figure \ref{fig:dvt_overview}, we align supervised samples belonging to the same class into the same Gaussian component in latent space, independently from their domain membership. This is possible thanks to  \authorimg{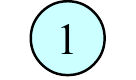}, an encoder $q_{\phi}(\mathbf{z}|\mathbf{x})$ composed of shared and specialized layers. The first generic layers are the same for all domains and learn generic features useful to classify in all domains. The last specialized layers are instead specific to each domain and are meant to compensate for covariate shift across domains. Once in latent space, samples with different domains but same class are aligned by forcing them into the same Gaussian mixture model. In the meanwhile, supervised samples allow for training a classifier $q_{\phi}(y|\mathbf{z})$ \authorimg{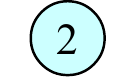}, later used to also label unsupervised data points from the latent space. When samples are unsupervised, we take the output of the classifier, logits, and use it to sample a class through Straight-Through Gumbel-Softmax \cite{jang2016categorical}. The sampled class tells us to which Gaussian mixture component we want to align the unsupervised sample. Finally, we deploy \authorimg{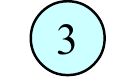}, a decoder modelling $p_{\phi}(\mathbf{x}|\mathbf{z})$ with a structure mirroring the encoder's structure. The decoder generates a sample, which probability is computed based on the original input samples as in any variational autoencoder.

Intuitively, we expect the reconstruction term to force the representation of supervised and unsupervised samples close in input space to be close in latent space too. In other terms, we expect that by forcing supervised samples toward their pre-determined Gaussian mixture component, the unsupervised samples close in input space will follow along, ultimately getting the unsupervised samples to fall into the right Gaussian component and being correctly classified. This allows for discovering an optimal latent representation to perform semi-supervised learning, while mitigating the risk to loose information about minority classes even when the model's capacity is not large enough.

As previously presented, past solutions often failed dealing with two challenges. First, working in unsupervised target conditions makes the problem of adaptation under full joint distributions changes $p_{S}(x,y) \neq p_{T}(x,y)$ intractable. The only solution is to force further assumptions not always realistic and incurring in negative transfer when applied to real datasets. Second, the majority of existing methods do not deal with imbalanced class distributions, failing in applications like our stars problem. DVT mitigates both problems by means of very few supervised samples from both source and target domains, combining the power of semi-supervised and transfer learning into a single framework. 

%\textbf{WW: ALL of the derivation in the following should be moved to an appendix, you should state only the objectives and the various approximations and reparametrizations.}

\subsection{SEMI-SUPERVISED LEARNING} \label{method-semi-supervised}

We first consider working on a single domain, where pairs $\mathcal{D} = \{ ( \mathbf{x}_{1} , y_{1} ),...,( \mathbf{x}_{N} , y_{N} ) \}$ represent our data. Each $i$-th pair is composed by an observation $\mathbf{x}_{i} \in \mathbb{R}^{D}$ and corresponding class label $y_{i} \in \left \{1,..., K\right \}$, $K$ equal to the total number of classes. Each observation has a corresponding latent variable $\mathbf{z}_{i}$. Due to the semi-supervised regime, only a subset of the observations will have a corresponding label $y_{i}$. When missing, we treat the label as a second latent variable. From now on we omit the index $i$, whenever it is clear that we work with single observations. With $\tilde{p}_{l}(\mathbf{x},y)$ and $\tilde{p}_{u}(\mathbf{x})$ we represent the empirical distribution over the labeled and unlabeled data respectively. We now describe DVT generative and inference models for a single domain.

\subsubsection{Generative model}

We propose a probabilistic model that describes the data as being generated from a latent Gaussian Mixture. More in detail, an observed sample $\mathbf{x}$ results from the following generative process:

\begin{equation}
\setlength\abovedisplayskip{0pt}
    \centering
    \displaystyle
      \begin{aligned} 
      p(y) &= \textrm{Cat}(y|\boldsymbol{\pi}); \\
	  p(\mathbf{z}|y) &= \mathcal{N}(\mathbf{z}|\boldsymbol{\mu}_{y}, \boldsymbol{\sigma}_{y}^{2}\textrm{I}); \\
		 p_{\theta}(\mathbf{x}|\mathbf{z}) &= f(\mathbf{x};\mathbf{z},\boldsymbol{\theta}), 
      \end{aligned}
\end{equation}

where $\pi_{y}$ is the prior probability for class $y$, $\boldsymbol{\pi} \in \mathbb{R}^{K}_{+}$, $1 = \sum_{y=1}^{K} \pi_{y}$, and $\textrm{Cat}(y|\pi)$ is a multinomial distribution parametrized by $\boldsymbol{\pi}$. $\boldsymbol{\mu}_{y}$ and $\boldsymbol{\sigma}_{y}^{2}$ are the mean and variance of the isotropic Gaussian distribution corresponding to class $y$, $\mathbf{I}$ is an identity matrix. $f(\mathbf{x};\mathbf{z},\boldsymbol{\theta})$ is a suitable likelihood function parametrized by a non-linear transformation of the latent variable $\mathbf{z}$ with parameters $\boldsymbol{\theta}$. We use a neural network as function approximator. If $\mathbf{x}$ is binary, the network outputs the parameters $\boldsymbol{\mu}_{x}$ of a Bernoulli distribution $\textrm{Ber}(\boldsymbol{\mu}_{x})$. If $\mathbf{x}$ is continuous, the output are the parameters $[\boldsymbol{\mu}_{x}; \textrm{log}\, \boldsymbol{\sigma}_{x}^{2}]$ of a Gaussian distribution $\mathcal{N}(\boldsymbol{\mu}_{x}, \boldsymbol{\sigma}_{x}^{2}\textrm{I})$.

\subsubsection{Recognition model}

The computation of the exact posterior distribution is intractable due to the non-linear and non-conjugate dependencies between the random variables. To solve the problem, we exploit advances in approximate Bayesian inference \cite{kingma2013auto,rezende2014stochastic}, introducing a fixed form distribution $q_{\phi}(y,\mathbf{z}|\mathbf{x})$ with global parameters $\boldsymbol{\phi}$ that approximate the true posterior distribution $p(y,\mathbf{z}|\mathbf{x})$.We assume $q_{\phi}(y,\mathbf{z}|\mathbf{x})$ factorizes as follows:

\begin{equation}
\setlength\abovedisplayskip{0pt}
   \centering
   \displaystyle
    \begin{aligned} 
      q_{\phi}(\mathbf{z}|\mathbf{x}) &= \mathcal{N}(\mathbf{z}|\boldsymbol{\mu}_{\phi}(\mathbf{x}),\boldsymbol{\sigma}_{\phi}^{2}(\mathbf{x})\mathbf{I})); \\
      q_{\phi}(y|\mathbf{z}) &= \textrm{Cat}(y|\boldsymbol{\pi}_{\phi}(\mathbf{z})), 
    \end{aligned}
\end{equation}

where we use a neural network to output means $\boldsymbol{\mu}_{\phi}(\mathbf{x})$ and standard deviations $\boldsymbol{\sigma}_{\phi}(\mathbf{x})$ to sample $\mathbf{z}$ from a normal distribution. Another network, that takes the form of a classifier, gets $\mathbf{z}$ as input and output the parameters to sample from a categorical distribution.

\subsubsection{Semi-supervised objective}

For labeled data, we perform inference on $z \sim q(\mathbf{z}|\mathbf{x})$ only. The variational lower bound on labeled data is given by:

\begin{equation} \label{loss_sup}
\setlength\abovedisplayskip{0pt}
%{\centering
   \displaystyle
    \begin{aligned}
      \textrm{log} \, p_{\theta}(\mathbf{x}) & \geq \mathbb{E}_{q_{\phi}(\mathbf{z}|\mathbf{x})} \left [ \textrm{log} \, \frac{p_{\theta}(\mathbf{x},\mathbf{z})}{q_{\phi}(\mathbf{z}|\mathbf{x})} \right ]  \\ 
	  &= \mathbb{E}_{q_{\phi}(\mathbf{z}|\mathbf{x})} \left [\textrm{log} \, p_{\theta}(\mathbf{x}|\mathbf{z}) - \textrm{log} \, \frac{q_{\phi}(\mathbf{z}|\mathbf{x})}{p_{\theta}(\mathbf{z})} \right ] \\
	  &= - \mathcal{L}^{sup}(\boldsymbol{\theta},\boldsymbol{\phi};\mathbf{x}).\\
    \end{aligned}  
%\par}
\end{equation}

The supervised ELBO looks identical to the one of vanilla Variational Autoencoders \cite{kingma2013auto,rezende2014stochastic} and the observed label $y$ never appears in the equation. However, the supervision comes in when computing the Kullback-Leibler divergence between $q_{\phi}(\mathbf{z}|\mathbf{x})$ and $p_{\theta}(\mathbf{z})$. In fact, we force $p_{\theta}(\mathbf{z})$ to be equal to the Gaussian belonging to the observed class $y$ with distribution $\mathcal{N}(\boldsymbol{\mu}_{y},\boldsymbol{\sigma}_{y}^{2}\mathbf{I})$.

When the label is missing, we treat $y$ as a latent variable. We perform posterior inference using the following lower bound:

\begin{equation} \label{loss_unsup}
\setlength\abovedisplayskip{0pt}
%{\centering
   \displaystyle
    \begin{aligned} 
      \textrm{log} \, p_{\theta}(\mathbf{x}) & \geq \mathbb{E}_{q_{\phi}(\mathbf{z},y|\mathbf{x})} \left [ \textrm{log} \, \frac{p_{\theta}(\mathbf{x},y,\mathbf{z})}{q_{\phi}(\mathbf{z},y|\mathbf{x})} \right ]  \\ 
	  &= \mathbb{E}_{q_{\phi}(\mathbf{z},y|\mathbf{x})} \left [\textrm{log} \, p_{\theta}(\mathbf{x}|\mathbf{z}) - \textrm{log} \, \frac{q_{\phi}(\mathbf{z},y|\mathbf{x})}{p_{\theta}(y,\mathbf{z})} \right ] \\
	  &= - \mathcal{L}^{unsup}(\boldsymbol{\theta},\boldsymbol{\phi};\mathbf{x}).\\
    \end{aligned}  
%\par}
\end{equation}

Combining the terms in equations \ref{loss_sup} and \ref{loss_unsup}, we derive the following bound on the marginal likelihood for the entire dataset:

\begin{equation} \label{eq:loss_j}
\setlength\abovedisplayskip{0pt}
%{\centering
   \displaystyle
    \begin{aligned} 
      \mathcal{J} = (1-\gamma) \sum_{(\mathbf{x},y) \sim \tilde{p}_{l}} \mathcal{L}^{sup} + \gamma \sum_{(\mathbf{x}) \sim \tilde{p}_{u}} \mathcal{L}^{unsup}
    \end{aligned}
%\par}
\end{equation}

The constant $\gamma$ introduced in eq. \ref{eq:loss_j} controls the relative strength of the supervised and unsupervised terms. This is important in situations where the difference in relative size between supervised and unsupervised sets is pronounced. Reading eq. \ref{eq:loss_j}, we observe that the label predictive distribution $q_{\phi}(y|\mathbf{x})$ does not contribute to the first labelled term. Instead, we would like to use labelled data to train the classifier. We remedy by adding a classification loss such that the distribution $q_{\phi}(y|\mathbf{x})$ learns from labelled data too (as in \cite{kingma2014semi}). The resulting extended loss function follows:

\begin{equation} \label{eq:loss_j_alpha}
\setlength\abovedisplayskip{0pt}
%{\centering
   \displaystyle
    \begin{aligned} 
      \mathcal{J}^{\rho} = \mathcal{J} + \rho \, \mathbb{E}_{\tilde{p}_{l}(\mathbf{x},y)} \left [ - \textrm{log} \, q_{\phi}(y|\mathbf{x}) \right ]
    \end{aligned}
%\par}
\end{equation}

\subsubsection{Semi-supervised transfer objective}

We can train DVT in two possible ways. On one hand, we can take a DVT model trained on source data, fix the shared layers and transfer knowledge by training the target specific layers only. On the other hand, we can train source and target domains simultaneously in a multi-task fashion. This is possible by extending equ. \ref{eq:loss_j_alpha} to average over source and target losses as follows:

\begin{equation} \label{eq:loss_j_alpha_transfer}
\setlength\abovedisplayskip{0pt}
%{\centering
   \displaystyle
    \begin{aligned} 
      \mathcal{J}^{DVT} = \eta \, \mathcal{J}^{S} + (1-\eta) \, \mathcal{J}^{T}.
    \end{aligned}
%\par}
\end{equation}

\subsubsection{Training}

$\boldsymbol{\mathcal{L}^{sup}}$ in eq. \ref{loss_sup} is a composition of two terms. The first log-likelihood expectation term,  $\mathbb{E}_{q_{\phi}(\mathbf{z}|\mathbf{x})} \left [\textrm{log} \, p_{\theta}(\mathbf{x}|\mathbf{z}) \right ]$, can be rewritten using the location-scale transformation for the Gaussian distribution as in \cite{kingma2013auto,rezende2014stochastic}:

\begin{equation} \label{likelihood_rep}
\setlength\abovedisplayskip{0pt}
%{\centering
   \displaystyle
    \begin{aligned} 
      \mathbb{E}_{\mathcal{N}(\boldsymbol{\epsilon}|\mathbf{0},\mathbf{\textrm{I}})} \left [\textrm{log} \, p_{\theta}(\mathbf{x}|\boldsymbol{\mu}_{\phi}(\mathbf{x}) + \boldsymbol{\sigma}^{2}_{\phi}(\mathbf{x}) \odot \boldsymbol{\epsilon} )\right ], \\
    \end{aligned}
%\par}
\end{equation}

where $\odot$ denotes an element-wise product. While the gradients of the loss in eq. \ref{likelihood_rep} cannot be computed analytically, gradients with respect to the generative and variational parameters $ \left \{ \phi, \theta \right \}$ can be computed as expectations of simple gradients as in \cite{kingma2013auto, rezende2014stochastic}:

\begin{equation} \label{}
\setlength\abovedisplayskip{0pt}
%{\centering
   \displaystyle
    \begin{aligned} 
      \mathbb{E}_{\mathcal{N}(\boldsymbol{\epsilon}|\mathbf{0},\mathbf{\textrm{I}})} \left [ \nabla_{\left \{ \boldsymbol{\theta}, \boldsymbol{\phi} \right \}} \textrm{log} \, p_{\theta}(\mathbf{x}|\boldsymbol{\mu}_{\phi}(\mathbf{x}) + \boldsymbol{\sigma}^{2}_{\phi}(\mathbf{x}) \odot \boldsymbol{\epsilon} )\right ]. \\
    \end{aligned}
%\par}
\end{equation}

The second term, $\mathbb{E}_{q_{\phi}(\mathbf{z}|\mathbf{x})} \left [ \textrm{log} \, \frac{q_{\phi}(\mathbf{z}|\mathbf{x})}{p_{\theta}(\mathbf{z})} \right ]$, is a KL divergence between two Gaussian distributions with an analytic solution.

$\boldsymbol{\mathcal{L}^{unsup}}$ in eq. \ref{loss_unsup} can be reduced to three terms: 

\begin{equation} \label{loss_unsup_kl_simpl}
\setlength\abovedisplayskip{0pt}
%{\centering
   \displaystyle
    \begin{aligned} 
      \mathcal{L}^{unsup} = & \,\, \mathbb{E}_{q_{\phi}(\mathbf{z}|\mathbf{x})}  \left [\textrm{log} \, p_{\theta}(\mathbf{x}|\mathbf{z}) \right ] + \\
      & + \mathbb{E}_{q_{\phi}(\mathbf{z}|\mathbf{x})} \left [ D_{KL}(q(y|\mathbf{z})||p(y)) \right ] + \\
      & +  \mathbb{E}_{q_{\phi}(y|\mathbf{z})} \left [ D_{KL}(q(\mathbf{z}|\mathbf{x})||p(\mathbf{z}|y)) \right ]. \\
    \end{aligned}
%\par}
\end{equation}

The first log-likelihood expectation term is equivalent to the one already computed in $\mathcal{L}^{sup}$. The second term does not offer an analytic solution, but it is possible to compute gradients by applying the reparametrization trick. In fact, it is possible to derive the following expectations of simple gradients:

\begin{equation} \label{gumbel_conditional_reparametrization}
\setlength\abovedisplayskip{0pt}
\centering
    \displaystyle
    \begin{aligned} 
      & \nabla_{\phi} \mathbb{E}_{q_{\phi}(y|\mathbf{z})} \left [ D_{KL}(q(\mathbf{z}|\mathbf{x})||p(\mathbf{z}|y)) \right ] \\
	  & = \mathbb{E}_{Gumbel(g|0,1)} \left [ \nabla_{\phi} D_{KL}(q(\mathbf{z}|\mathbf{x})||p(\mathbf{z}|f(\phi, g, \tau)) \right ] , \\
	  & \quad \,\, \textrm{with} \; f_{k}(\phi,g,\tau)_{1 \leq k \leq K} = \frac{\textrm{exp} \, (( \alpha_{k} + g_{k} )/\tau))}{\sum_{i=1}^{K} \textrm{exp} \, (( \alpha_{i} + g_{i} )/\tau)} .
    \end{aligned}
\end{equation}

By means of the Gumbel-Max \cite{jang2016categorical, maddison2016concrete}, we are able to reparameterize the categorical distribution $q_{\phi}(y|\mathbf{z})$ as a deterministic function $f$ of $\phi$ and a random variable $g \sim Gumbel(0,1)$, where $g$ does not depend on $\phi$. However, Gumbel-Max definition comprehend an $arg\_max$ operation, which derivative is $0$ everywhere except at the boundary of state changes, where it is not defined. Therefore we use the Gumbel-Softmax relaxation \cite{jang2016categorical, maddison2016concrete} as defined in $f(\phi,g,\tau)$ in order to take gradients of $D_{KL}$ with respect to $\phi$. Due to the relaxation, $p(\mathbf{z}|f(\phi, g, \tau)$ is a mixture of Gaussians. In order to get a Kullback-Leibler divergence between two gaussian distributions, allowing us to retrieve an analytic solution, we use the Straight-Through Gumbel estimator as defined in \cite{jang2016categorical}, which discretize $y$ using $arg\_max$ during forward pass, while using the continuous approximation in the backward pass. The final expectation over gradients follows: 

\begin{equation} \label{gumbel_conditional_max}
\setlength\abovedisplayskip{0pt}
\centering
    \displaystyle
    \begin{aligned} 
    	  \mathbb{E}_{Gumbel(g|0,1)} & \left [ \nabla_{\phi}  D_{KL}(\mathcal{N}(\boldsymbol{\mu}_{\phi}, \boldsymbol{\sigma}^{2}_{\phi}\mathbf{I})||\mathcal{N}(\boldsymbol{\mu}_{y}, \boldsymbol{\sigma}^{2}_{y}\mathbf{I}) \right ] , \\
    \end{aligned} 
\end{equation}

where $y$ is equal to $\textrm{arg\_max}(f(\phi, g, \tau))$. While this makes our estimator a biased one, we see it works well in practice. Following the same procedure, we can derive an expectation over gradients for the third term in eq. \ref{loss_unsup_kl_simpl}:

\begin{equation} \label{gumbel_prior_reparametrization}
\setlength\abovedisplayskip{0pt}
\centering
    \displaystyle
    \begin{aligned} 
      & \nabla_{\phi} \mathbb{E}_{q_{\phi}(\mathbf{z}|\mathbf{x})} \left [ D_{KL}(q(y|\mathbf{z})||p(y))\right ] \\
      & = \mathbb{E}_{\mathcal{N}(\epsilon|0,1)} \left [ \nabla_{\phi} D_{KL}(\boldsymbol{\pi}_{\phi}(f(\mathbf{x},\boldsymbol{\epsilon}))||p(y)) \right ] , \\
      & \quad\,\, \textrm{with} \; f(\mathbf{x},\boldsymbol{\epsilon}) = \boldsymbol{\mu}_{\phi}(\mathbf{x}) + \boldsymbol{\sigma}^{2}_{\phi}(\mathbf{x}) \odot \boldsymbol{\epsilon}.
      \end{aligned}
\end{equation}

For more details on the various derivations, please refer to the additional material.

\begin{table}%[!ht]
    \centering
    \begin{tabular}{|c|c|}
        \hline
        \textbf{Parameters} & \textbf{Value} \\ \hline
        $\lambda,\rho$ & $0.1,10^{4}$ \\ \hline
        $\mu,\sigma$ & $10,0.1$ \\ \hline
        $\eta,\beta_{1},\beta_{2},\epsilon$ & $0.005,0.5,0.5,0.001$ \\ \hline
    \end{tabular}
    \caption{Hyper-parameters used across all of the experiments.}
    \label{hyper-parameters}
\end{table}

\begin{figure}%[!ht]
    \begin{subfigure}{\linewidth}
        \begin{subfigure}{.48\linewidth}
            \caption*{Predictions}
            \includegraphics[width=\linewidth]{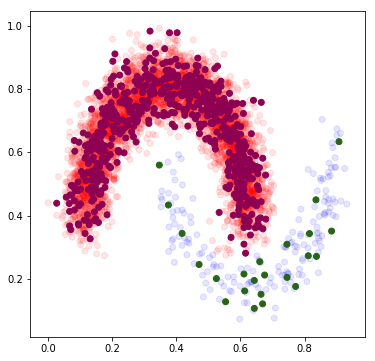}
        \end{subfigure}
        \begin{subfigure}{.48\linewidth}
            \caption*{Entropy}
            \includegraphics[width=\linewidth]{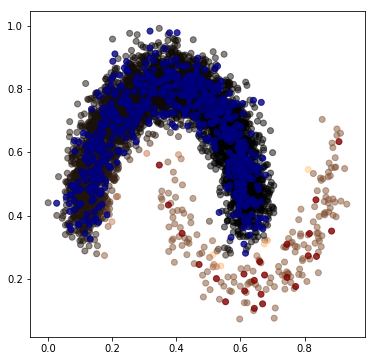}
        \end{subfigure}
        \caption{DVT after semi-supervised training on source dataset.}
    \end{subfigure}
    \begin{subfigure}{\linewidth}
        \begin{subfigure}{.48\linewidth}
            \includegraphics[width=\linewidth]{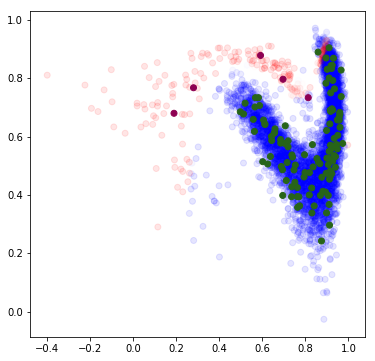}
        \end{subfigure}
        \begin{subfigure}{.48\linewidth}
            \includegraphics[width=\linewidth]{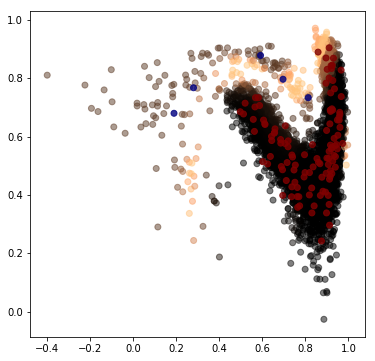}
        \end{subfigure}
        \caption{DVT after transfer on target dataset.}
    \end{subfigure}
\caption{A variation of the inter-twinning moons problem. In the left column, target samples predictions are represented as transparent red and blue dots. Correspondent supervised training samples are represented as thick violet (red class) and green (blue class) dots. In the right column, we plot the target predictions' binary entropy: the brighter the color, the higher the uncertainty.}
\label{fig:synthetic_experiment}
\end{figure}

\section{EXPERIMENTS}

%\textbf{WW: Get rid of all mention of multi-task learning! move all tables summarizing datasets and parameters into appendix!}

In this section, we present the experiments performed on synthetic, digits and real stars datasets.

\subsection{SYNTHETIC} \label{ssec:synthetic-experiment}

%\begin{figure}%[!ht]
%\centering
%\begin{subfigure}[b]{.48\linewidth}
%\includegraphics[width=\linewidth]{original_dataset.pdf}
%\caption{Source dataset.}\label{fig:original_dataset}
%\end{subfigure}
%\begin{subfigure}[b]{.48\linewidth}
%\includegraphics[width=\linewidth]{rotated_dataset.pdf}
%\caption{Target dataset.}\label{fig:rotated_dataset}
%\end{subfigure}
%\caption{Synthetic datasets generated to test DVT.}
%\label{fig:synthetic_datasets}
%end{figure}

In our first experiment, we study DVT's performances on a synthetic problem: a variant of the inter-twinning 2D moons \cite{ganin2016domain} presenting generic dataset shift.

%Figure \ref{fig:synthetic_datasets} presents the generated source and target datasets. Both $p(x|y)$ and $p(y)$ are changing at the same time, making this synthetic task both a challenging one and a representative failure example for most unsupervised transfer learning methods. 

%\textbf{[WW: You get ONE SHORT paragraph to describe the setup of this experiment. Move all parameter settings and numerical distractions to the APPENDIX]}

\textbf{Data.} We generate an unbalanced source dataset composed by two inter-twinning moons, with $10000$ samples for class red and $400$ for class blue. We split the dataset in half to obtain train and test dataset. We then randomly sample $\sim 10\%$ training examples to be supervised, the remaining as unsupervised. We repeat the same procedure for target data, but with reverted class balance and applying a two step transformation to $p_{T}(x)$: (1) rotate by $30^{\circ}$ with respect to the center $[.5,.5]$; (2) apply a logarithmic transformation $f(x) = \textrm{log}_{10} \, (x) + 1$. This time only $\sim 2.5\%$ of the samples are considered supervised, thus having only five examples from the (red) minority target class. See figure \ref{fig:synthetic_experiment} to visualize the transformation.

\textbf{Hyper-parameters.} We select hyper-parameters based on semi-supervised performances on the source domain. We perform a random search and select the model with best unweighted average F1 score with 5-fold cross-validation. Finally, we select the robust combination in table \ref{hyper-parameters} as the set of hyper-parameters for all the experiments. For further details on the hyper-parameters selection and the tested ranges, please refer to the supplementary material.

\textbf{Architecture.} We used an encoder with three fully connected layers. In the order, two shared layers and one domain specific layer. The decoder mirrors the encoder structure. We used single layer networks as domain specific classifiers. Refer to the supplementary material for a summary table.

\textbf{Train.} After training on source data, we freeze the shared layers and train the target specific layers only. In both phases, we use a supervised batch equal to the number of supervised samples and 100-sized batches for unsupervised data. We stop the training after $15000$ batches.

\textbf{Evaluation.} As for hyper-parameter selection, we used unweighted mean F1 score as to equally balance the importance of all classes, independently from their sizes.

%\begin{figure}
%  \centering
%  \includegraphics[width=.8\linewidth]{mu_setting.pdf}
%  \caption{Picture showing how we build the latent Gaussian mixture prior. In particular, the picture shows how we place two Gaussian components, with identical diagonal covariance matrix, in a three dimensional latent space, as a way to accommodate the synthetic binary classification problem.}
%  \label{fig:mu_setting}
%\end{figure}

\textbf{Results.} The first row of figure \ref{fig:synthetic_experiment} shows the semi-supervised learning results on the source dataset. On the left, we see how the two half moons are well classified with a $97.5\%$ unweighted mean F1 score. On the right, the binary cross-entropy allows us to observe how the uncertainty is particularly concentrated on the decision boundary and more in general on the minority class.

The second row of figure \ref{fig:synthetic_experiment} shows the transfer learning results on the target dataset. On the left, we observe how DVT is able to correctly classify almost all target samples with a $86.6\%$ unweighted mean F1 score, even if the number of provided supervised examples for the minority (red) class is reduced to $5$. On the right, we see how the decision boundary is well transferred, with potential applications to active learning strategies. Even more interestingly, the minority class is not considered uncertain in its entirety, as in the semi-supervised source case. In fact, red samples above the decision boundary had been classified with high certainty into the minority target class. This certainty is reminiscent of the fact that the red class in the source dataset represents the majority class instead of the minority one.

\subsection{DIGITS} \label{ssec:digits-experiment}

\begin{table}%[!ht]
\centering
\begin{tabular}{|r|c|c|}
\hline
\textbf{Training} & \multicolumn{1}{l|}{\textbf{Model}} & \multicolumn{1}{c|}{\textbf{Accuracy}} \\ \hline
\textit{Supervised} & \textit{DVT} & \textit{98.71 $\pm$ 0.04} \\
\textit{Supervised} & \textit{CNN} & \textit{98.58 $\pm$ 0.03} \\ \hline
Semi-supervised & DVT & \textbf{97.64 $\pm$ 0.33} \\
Transfer & DVT & 96.10 $\pm$ 0.21 \\
Multi-task & DVT & 96.13 $\pm$ 0.26 \\
Transfer & CNN & 93.67 $\pm$ 0.64 \\
Multi-task & CNN & 67.79 $\pm$ 2.06 \\
\hline
\end{tabular}
\caption{MNIST average accuracy $\pm$ standard deviation.}
\label{mnist-performances}
\end{table}

\begin{table}%[!ht]
\centering
\begin{tabular}{|r|c|c|}
\hline
\textbf{Training} & \multicolumn{1}{l|}{\textbf{Model}} & \multicolumn{1}{c|}{\textbf{Accuracy}} \\ \hline
\textit{Supervised} & \textit{DVT} & \textit{96.05 $\pm$ 0.06} \\
\textit{Supervised} & \textit{CNN} & \textit{95.89 $\pm$ 0.05} \\ \hline
Semi-supervised & DVT & 92.03 $\pm$ 0.38 \\
Transfer & DVT & \textbf{94.37 $\pm$ 0.19} \\
Multi-task & DVT & 92.13 $\pm$ 0.30 \\
Transfer & CNN & 88.26 $\pm$ 0.71 \\
Multi-task & CNN & 75.05 $\pm$ 1.82 \\ \hline
\end{tabular}
\caption{USPS average accuracy $\pm$ standard deviation.}
\label{usps-performances}
\end{table}

\begin{table}%[!ht]
    \centering
    \resizebox{.93\linewidth}{!}{
    \begin{tabular}{r|c|c|c|c|c|}
    \cline{2-6}
    \multicolumn{1}{l|}{} & \multicolumn{5}{c|}{\textbf{Classes}} \\ \hline
    \multicolumn{1}{|r|}{\textbf{Dataset}} & \textbf{CEP} & \textbf{EB} & \multicolumn{1}{l|}{\textbf{LPV}} & \multicolumn{1}{l|}{\textbf{QSO}} & \multicolumn{1}{l|}{\textbf{RRLYR}} \\ \hline
    \multicolumn{1}{|r|}{Eros} & 1.82 & 46.37 & 5.65 & 0.83 & 45.33 \\
    \multicolumn{1}{|r|}{Macho} & 1.27 & 18.75 & 19.66 & 4.98 & 55.34 \\
    \multicolumn{1}{|r|}{Hits} & 0. & 27.60 & 0. & 6.39 & 66.01 \\ \hline
    \end{tabular}
    }
    \caption{Differences in class distribution $p(y)$. Values expressed in percentages of the total samples per dataset.}
    \label{dataset-p-y}
\end{table}

In our second experiment, we study DVT's performances on images as a way to test its ability to reach high performances even when working with high-dimensional and heterogeneous data. More in detail, we used MNIST and USPS, two digits datasets presenting different input space dimensionalities.

\begin{table*}
\centering
\begin{tabular}{ll|c|c|c|c|c|}
\cline{3-7}
 &  & \multicolumn{5}{c|}{\textbf{Classes}} \\ \hline
\multicolumn{1}{|l|}{\textbf{Training}} & \textbf{Models} & CEP & EB & LPV & QSO & RRLYR \\ \hline
\multicolumn{1}{|l|}{\textit{Supervised upper bound}} & \textit{Random Forest} & \textit{80} & \textit{88} & \textit{97} & \textit{83} & \textit{95} \\ 
\multicolumn{1}{|l|}{\textit{Supervised upper bound}} & \textit{DVT} & \textit{80} & \textit{87} & \textit{98} & \textit{81} & \textit{94} \\ \hline
\multicolumn{1}{|l|}{Transfer \cite{benavente2017automatic}} & GMM based & 33 & 73 & 93 & 37 & 91 \\ 
\multicolumn{1}{|l|}{Multi-task top cycle consistency} & DVT & 44 $\pm$ 23 & 76 $\pm$ 5.2 & 95 $\pm$ 3.0 & 77 $\pm$ 2.4 &  93 $\pm$ 1.1 \\ 
\multicolumn{1}{|l|}{\textbf{Multi-task ensemble}} & \textbf{DVT} & \textbf{56} $\pm$ \textbf{26} & \textbf{85} $\pm$ \textbf{2.2} & \textbf{98} $\pm$ \textbf{0.4} & \textbf{84} $\pm$ \textbf{1.3} & \textbf{95} $\pm$ \textbf{0.6} \\ \hline
\end{tabular}
\caption{Performances Eros to Macho expressed in average percentage $\pm$ standard deviation when available.}
\label{eros-to-macho}
\end{table*}

\begin{table*}
\centering
\begin{tabular}{ll|c|c|c|c|c|}
\cline{3-7}
 &  & \multicolumn{5}{c|}{\textbf{Classes}} \\ \hline
\multicolumn{1}{|l|}{\textbf{Training}} & \textbf{Models} & CEP & EB & LPV & QSO & RRLYR \\ \hline
\multicolumn{1}{|l|}{\textit{Supervised upper bound}} & \textit{Random Forest} & \textit{86} & \textit{92} & \textit{93} & \textit{40} & \textit{93} \\ 
\multicolumn{1}{|l|}{\textit{Supervised upper bound}} & \textit{DVT} & \textit{86} & \textit{89} & \textit{92} & \textit{64} & \textit{89} \\ \hline
\multicolumn{1}{|l|}{Transfer \cite{benavente2017automatic}} & GMM based & 25 & 76 & 58 & 2 & 69 \\
\multicolumn{1}{|l|}{Multi-task top cycle consistency} & DVT & 73 $\pm$ 1.7 & 84 $\pm$ 0.5 & 83 $\pm$ 1.9 & 17 $\pm$ 9.1 &  84 $\pm$ 0.7 \\ 
\multicolumn{1}{|l|}{\textbf{Multi-task ensemble}} & \textbf{DVT} & \textbf{75} $\pm$ \textbf{0.9} & \textbf{87} $\pm$ \textbf{0.4} & \textbf{86} $\pm$ \textbf{0.2} & \textbf{18} $\pm$ \textbf{10} & \textbf{87} $\pm$ \textbf{0.5} \\ \hline
\end{tabular}
\caption{Performances Macho to Eros expressed in average percentage $\pm$ standard deviation when available.}
\label{macho-to-eros}
\end{table*}

\begin{table*}
\centering
\begin{tabular}{ll|c|c|c|}
\cline{3-5}
 &  & \multicolumn{3}{c|}{\textbf{Classes}} \\ \hline
\multicolumn{1}{|l|}{\textbf{Training}} & \textbf{Models} & EB & QSO & RRLYR \\ \hline
\multicolumn{1}{|l|}{\textit{Supervised upper bound}} & \textit{Random Forest} & \textit{93} & \textit{94} & \textit{98} \\ 
\multicolumn{1}{|l|}{\textit{Supervised upper bound}} & \textit{DVT} & \textit{93} & \textit{92} & \textit{98} \\ \hline
\multicolumn{1}{|l|}{Transfer \cite{benavente2017automatic}} & GMM based & 2 & 37 & 15 \\
\multicolumn{1}{|l|}{Multi-task top cycle consistency} & DVT & 85 $\pm$ 3.4 & 83 $\pm$ 6.2 & 94 $\pm$ 0.4 \\
\multicolumn{1}{|l|}{\textbf{Multi-task ensemble}} & \textbf{DVT} & \textbf{88} $\pm$ \textbf{1.7} & \textbf{88} $\pm$ \textbf{2.7} & \textbf{95} $\pm$ \textbf{0.9} \\ \hline
\end{tabular}
\caption{Performances Eros to Hits expressed in average percentage $\pm$ standard deviation when available.}
\label{eros-to-hits}
\end{table*}

\textbf{Data.} The MNIST dataset \cite{lecun1998gradient} consists of $60000$ examples of $0$ - $9$ handwritten digits for training and $10000$ digits for testing. The images are centered and of size $28$ by $28$ pixels. The USPS dataset \cite{hull1994database} contains $7291$ examples of $0$ - $9$ postal code digits for training and $2007$ examples for testing. The images are centered and of size $16$ by $16$ pixels. This dataset is considered quite hard with reported human error rate of $2.5\%$.

\textbf{Architecture.} We used an encoder with five convolutional layers. In the order, three shared layers and two domain specific layers. The decoder mirrors the encoder structure, a part from the last layers being domain specific as to deal with different output dimensionalities. We used a single and shared fully-connected layer as domain classifier, being the class distribution unchanged across domains. We set up a Convolutional Neural Network (CNN) to compare DVT against classic techniques. To have a fair comparison, we set the CNN's architecture to be the concatenation of DVT's encoder and classifier without $\mathbf{z}$ and $y$ sampling operations. For details, refer to the supplementary material.

\textbf{Train.} We train DVT with three different strategies: semi-supervised, transfer and multi-task learning. The latter is performed by training simultaneously on source and target with equation \ref{eq:loss_j_alpha_transfer}, with $\eta=.5$, leaving all the parameters free to be trained at once. For all three strategies, we sample $100$ training examples to be used as supervised data points, while the remaining train set is considered as unsupervised. We train the CNN in two ways: (i) transfer mode by training on source and retraining last three layers on target; (ii) multi-task mode by training a CNN by, at each back-propagation, equally weighting the loss computed on each single batch per domain.

\textbf{Evaluation.} We evaluate all models in terms of accuracy on test data.

\textbf{Results.} Table \ref{mnist-performances} reports the results for the USPS to MNIST experiments. DVT, trained in a semi-supervised learning fashion with as few as $100$ labels, almost reaches the upper-bound supervised performances of both DVT and CNN models. Semi-supervised learning on MNIST attains better mean accuracy than both transfer and multi-task learning. Yet, both transfer and multi-task keep being competitive and not more than $1.6\%$ away from their semi-supervised counterpart. We attribute the inability of DVT to perform in transfer and multi-task mode as well as in semi-supervised mode on MNIST to the low-resolution and smaller size of USPS. Notably, we observe a striking difference in performance between DVT and CNN when both trained in multi-task mode with as few as $100$ labels each: DVT attains almost $+30\%$ in accuracy. DVT shows the possibility to train deep networks even when very few labelled points are available.

DVT reaches better accuracy performances against a number of deep semi-supervised techniques: $+0.97\%$ over full M1+M2 model presented in \cite{kingma2014semi} and $+4.04\%$ against \cite{jang2016categorical}. DVT still performs worse than the state-of-the-art ladder model presented in \cite{rasmus2015semi} with a $-1.79\%$, however the "low" accuracy of the fully supervised DVT let us think that the gap might be reduced with computationally intensive hyper-parameters' space search.

Table \ref{usps-performances} reports the results for the MNIST to USPS experiments. In this case we see how transferring reach information from a higher resolution and more varied dataset such as MNIST helps improving learning performances on USPS. In particular, we observe how DVT transfer achieves a $+2.34\%$ in accuracy against its semi-supervised counterpart. Once again, we note how when both DVT and CNN are presented with very few supervised labeles in multi-task mode, DVT achieves much better performances with an almost $+20\%$ improvement.

\subsection{STARS} \label{ssec:stars-experiment}

In our third experiment, we consider a real world problem: transfer learning across stars datasets. This task is particularly challenging due to the simultaneous presence of high class imbalance, changes in class distribution and shifts in class conditional distribution.

\textbf{Data.} We consider three datasets containing stars light curves: EROS \cite{beaulieu1994spectroscopic}, MACHO \cite{alcock1997macho} and HiTS \cite{forster2016high}. We use the Feature Analysis for Time Series (FATS) Python package to automatically extract features from the light curves \cite{nun2015fats}. We apply further pre-processing steps on HiTS, as presented in \cite{martinez2018high}. Table \ref{dataset-p-y} summarizes the class distributions across the datasets under analysis, showing clear differences in $p(y)$. \cite{benavente2017automatic} also demonstrates the presence of conditional-shifts. When considering a dataset as a target domain,  and sample $0.1\%$, $1\%$ and $5\%$ from EROS, MACHO and HiTS respectively as supervised set. The difference in percentage adjusts for the differences in the number of samples available in the three datasets, in the order $26005$, $1104$ and $924$.

\textbf{Architecture.} We used an encoder with three fully connected layers: two shared and one domain specific. The decoder mirrors the encoder structure. We used a single domain specific fully-connected layer as domain classifier. Please, refer to supplementary material for further details.

\textbf{Train.} We found multi-task training to be particularly successful, clearly outperforming transfer training. We therefore selected multi-task mode as training procedure for this experiment. After noticing an over-fitting problem, we decide to force a training early-stop event when noticing a decreasing cycle consistency effect. Let us consider $G$ as encoding function, $F$ as decoding function and $C$ as classification function. Cycle consistency is defined as the classification performance achieve after applying the following transformations to the abundant labelled source data: $C_{S}(G_{T}(F_{T}(G_{S}(\mathbf{x})))$. We repeat each experiment ten times in order to compute the standard deviation. At each time, we sample a supervised set of source and target examples and we train ten models.  In tables \ref{eros-to-macho}, \ref{macho-to-eros} and \ref{eros-to-hits} we report the performance statistics for top cycle consistency models as "Multi-task top cycle consistency" and average ensembles of all ten models as "Multi-task ensemble".

\textbf{Evaluation.} We used unweighted mean F1 score as in \cite{benavente2017automatic} on training unsupervised data points.

\textbf{Results.}  In the Eros to Macho experiment (table \ref{eros-to-macho}), we observe how DVT almost doubles classification performances on the rarest Macho class, CEP, even when provided with one single supervised CEP sample, passing from $33\%$ to $56\%$ F1 score. We also observe how, depending on the supervised sample provided, we might not have enough supervised guidance to model CEP stars, leading to occasional drops in performances and high variance results. However, this issue is strongly reduced as soon as we slightly increase the number of supervised samples. An example of this phenomenon is visible for QSO, for which 6 Macho supervised examples are provided. 

Further confirmations come from the Macho to Eros experiment where two supervised samples of the rarest class, QSO, are sufficient to highly reduce variance results and reach important F1 improvements. Namely we pass $2\%$ to $18\%$ F1 score. The variance reduction for extremely reduced increase in number of supervised samples is again confirmed by the CEP classification performances, where we pass from $25\%$ to $75\%$ F1 scores and very low performances' variance with just 5 supervised CEP samples from EROS. Finally, we observe in table \ref{eros-to-hits} how DVT performs better than \cite{benavente2017automatic} in all classes. Yet, these results are not comparable because, differently from us, Benavente's work \cite{benavente2017automatic} did not apply the cleaning procedure applied on HiTS data as presented in \cite{martinez2018high}.

\section{CONCLUSION}

We present DVT, a method to correct for generic dataset shifts. DVT is able to simultaneously: (i) perform semi-supervised learning on each single domain to maximally exploit both supervised and unsupervised samples; (ii) correct for domain shift by aligning samples from different domains in a single latent Gaussian mixture distribution. We apply DVT to synthetic samples, digits datasets and astronomical surveys. We demonstrate the DVT's ability not only to fit different datasets, but also different training procedures: semi-supervised, transfer and multi-task. Our results show that a significant performance gain in classification can be obtained using DVT, at the minimal cost of few target labeled data.

{\small
\bibliographystyle{ieee}
\bibliography{arxiv}
}

\pagebreak

%\section{Supplementary Material: model optimization}

\pagebreak
\begin{center}
\textbf{\large Supplementary material: model optimization}
\end{center}

$\boldsymbol{\mathcal{L}^{sup}}$ can be decomposed into $\mathbb{E}_{q_{\phi}(\mathbf{z}|\mathbf{x})} \left [\textrm{log} \, p_{\theta}(\mathbf{x}|\mathbf{z}) \right ] - \mathbb{E}_{q_{\phi}(\mathbf{z}|\mathbf{x})} \left [ \textrm{log} \, \frac{q_{\phi}(\mathbf{z}|\mathbf{x})}{p_{\theta}(\mathbf{z})} \right ]$, where the first log-likelihood expectation term can be rewritten using the location-scale transformation for the Gaussian distribution \cite{kingma2013auto,rezende2014stochastic} as

\begin{equation} \label{}
\setlength\abovedisplayskip{0pt}
%{\centering
   \displaystyle
    \begin{aligned} 
      \mathbb{E}_{\mathcal{N}(\boldsymbol{\epsilon}|\mathbf{0},\mathbf{\textrm{I}})} \left [\textrm{log} \, p_{\theta}(\mathbf{x}|\boldsymbol{\mu}_{\phi}(\mathbf{x}) + \boldsymbol{\sigma}^{2}_{\phi}(\mathbf{x}) \odot \boldsymbol{\epsilon} )\right ]. \\
    \end{aligned}
%\par}
\end{equation}

The second term is a KL divergence between two gaussians, that we can compute analytically following proof in \cite{jiang2016variational}:

\begin{equation} \label{kl_gaussians}
\setlength\abovedisplayskip{0pt}
%{\centering
   \displaystyle
    \begin{aligned} 
       & D_{KL}(\mathcal{N}(\tilde{\boldsymbol{\mu}}_{\phi},\tilde{\boldsymbol{\sigma}}_{\phi}^{2}\mathbf{I})||\mathcal{N}(\boldsymbol{\mu}_{y},\boldsymbol{\sigma}^{2}_{y}\mathbf{I})) = \frac{J}{2} \; \textrm{log} \; (2\pi) \\
       & + \frac{1}{2} ( \sum_{j=1}^{J} \; \textrm{log} \; \sigma_{yj}^{2} + \sum_{j=1}^{J} \frac{\tilde{\sigma}_{j}^{2}}{\sigma_{yj}^{2}} + \sum_{j=1}^{J} \frac{(\tilde{\mu}_{j} - \mu_{yj}^{2})}{\sigma_{yj}^{2}}). \\
    \end{aligned}
%\par}
\end{equation}

$\boldsymbol{\mathcal{L}^{unsup}}$ can decomposed into two terms. The first log-likelihood term can be rewritten using the same location-scale transformation used in $\mathcal{L}^{sup}$ once we notice the independence between $p_{\theta}(\mathbf{x}|\mathbf{z})$ and $\mathbb{E}_{q_{\phi}(y|\mathbf{z})}$:

\begin{equation} \label{rep_trick_unsup}
\setlength\abovedisplayskip{0pt}
%{\centering
   \displaystyle
    \begin{aligned} 
      &\mathbb{E}_{q_{\phi}(\mathbf{z},y|\mathbf{x})}  \left [\textrm{log} \, p_{\theta}(\mathbf{x}|\mathbf{z}) \right ] \\
      & \quad = \mathbb{E}_{q_{\phi}(\mathbf{z}|\mathbf{x})} \mathbb{E}_{q_{\phi}(y|\mathbf{z})} \left [\textrm{log} \, p_{\theta}(\mathbf{x}|\mathbf{z}) \right ] \\
      & \quad = \mathbb{E}_{q_{\phi}(\mathbf{z}|\mathbf{x})} \left [\textrm{log} \, p_{\theta}(\mathbf{x}|\mathbf{z}) \right ] \\
      & \quad = \mathbb{E}_{\mathcal{N}(\boldsymbol{\epsilon}|\mathbf{0},\mathbf{\textrm{I}})} \left [\textrm{log} \, p_{\theta}(\mathbf{x}|\boldsymbol{\mu}_{\phi}(\mathbf{x}) + \boldsymbol{\sigma}^{2}_{\phi}(\mathbf{x}) \odot \boldsymbol{\epsilon} )\right ] ,\\
    \end{aligned}
%\par}
\end{equation}

where $\odot$ denotes an element-wise product. While the gradients of the loss in equation \ref{rep_trick_unsup} cannot be computed analytically, gradients with respect to the generative and variational parameters $\nabla_{\left \{ \boldsymbol{\theta}, \boldsymbol{\phi} \right \}}  \mathbb{E}_{q_{\phi}(\mathbf{z}|\mathbf{x})} \left [\textrm{log} \, p_{\theta}(\mathbf{x}|\mathbf{z}) \right ]$ can be computed as expectations of simple gradients:

\begin{equation} \label{}
\setlength\abovedisplayskip{0pt}
%{\centering
   \displaystyle
    \begin{aligned} 
      \mathbb{E}_{\mathcal{N}(\boldsymbol{\epsilon}|\mathbf{0},\mathbf{\textrm{I}})} \left [ \nabla_{\left \{ \boldsymbol{\theta}, \boldsymbol{\phi} \right \}} \textrm{log} \, p_{\theta}(\mathbf{x}|\boldsymbol{\mu}_{\phi}(\mathbf{x}) + \boldsymbol{\sigma}^{2}_{\phi}(\mathbf{x}) \odot \boldsymbol{\epsilon} )\right ]. \\
    \end{aligned}
%\par}
\end{equation}

The second term can be further divided in two terms:

\begin{equation} \label{loss_unsup_kl}
  \setlength\abovedisplayskip{0pt}
   \displaystyle
    \begin{aligned} 
      & \mathbb{E}_{q_{\phi}(\mathbf{z},y|\mathbf{x})} \left [\textrm{log} \, \frac{q_{\phi}(\mathbf{z},y|\mathbf{x})}{p_{\theta}(y,\mathbf{z})} \right ]\\
      & = \mathbb{E}_{q_{\phi}(\mathbf{z},y|\mathbf{x})} \left [\textrm{log} \, \frac{q_{\phi}(\mathbf{z}|\mathbf{x})}{p_{\theta}(\mathbf{z}|y)} \right ] + \mathbb{E}_{q_{\phi}(\mathbf{z},y|\mathbf{x})} \left [\textrm{log} \, \frac{q_{\phi}(y|\mathbf{z})}{p_{\theta}(y)} \right ] \\
      & = \mathbb{E}_{q_{\phi}(y|\mathbf{z})} \mathbb{E}_{q_{\phi}(\mathbf{z}|\mathbf{x})} \left [\textrm{log} \, \frac{q_{\phi}(\mathbf{z}|\mathbf{x})}{p_{\theta}(\mathbf{z}|y)} \right ] + \\
      & \quad\quad\quad\quad\quad\quad\quad\quad\quad\quad \mathbb{E}_{q_{\phi}(y|\mathbf{z})} \mathbb{E}_{q_{\phi}(\mathbf{z}|\mathbf{x})} \left [\textrm{log} \, \frac{q_{\phi}(y|\mathbf{z})}{p_{\theta}(y)} \right ] \\
      & = \mathbb{E}_{q_{\phi}(y|\mathbf{z})} \left [ D_{KL}(q(\mathbf{z}|\mathbf{x})||p(\mathbf{z}|y)) \right ] + \\
      & \;\quad\quad\quad\quad\quad\quad\quad\quad\quad\quad \mathbb{E}_{q_{\phi}(\mathbf{z}|\mathbf{x})} \left [ D_{KL}(q(y|\mathbf{z})||p(y)) \right ]
    \end{aligned}
\end{equation}

In the last step, $\mathbb{E}_{q_{\phi}(y|\mathbf{z})} \left [ D_{KL}(q(\mathbf{z}|\mathbf{x})||p(\mathbf{z}|y)) \right ]$ can be trivially derived just noticing that the internal expectation is a Kullback-Leibler divergence. It is possible to derive the second term as follows:

\begin{equation} \label{kl_prior_derivation}
   \setlength\abovedisplayskip{0pt}
   \setlength\belowdisplayskip{0pt}
   \displaystyle
    \begin{aligned} 
      & \mathbb{E}_{q_{\phi}(y|\mathbf{z})} \mathbb{E}_{q_{\phi}(\mathbf{z}|\mathbf{x})} \left [\textrm{log} \, \frac{q_{\phi}(y|\mathbf{z})}{p_{\theta}(y)} \right ] \\
      & = \sum_{y}\int_{z} q(y|\mathbf{z}) q(\mathbf{z}|\mathbf{x}) \, \textrm{log} \, \frac{q(y|\mathbf{z})}{p(y)} dz \\
      & = \int_{z} q(\mathbf{z}|\mathbf{x}) \sum_{y} q(y|\mathbf{z}) \, \textrm{log} \, \frac{q(y|\mathbf{z})}{p(y)} dz \\
      & = \int_{z} q(\mathbf{z}|\mathbf{x}) D_{KL}(q(y|\mathbf{z})||p(y)) dz \\
      & = \mathbb{E}_{q_{\phi}(\mathbf{z}|\mathbf{x})} \left [ D_{KL}(q(y|\mathbf{z})||p(y)) \right ]
    \end{aligned}
\end{equation}

While it is not possible to write an analytical solution of the two expectations over KL divergences, it is possible to combine the expectations of simple gradients and the reparametrization trick to derive a simple Monte Carlo estimator. Here is the derivation for the first term:

\begin{equation} \label{gumbel_conditional_reparametrization_}
\setlength\abovedisplayskip{0pt}
\centering
    \displaystyle
    \begin{aligned} 
      & \nabla_{\phi} \mathbb{E}_{q_{\phi}(y|\mathbf{z})} \left [ D_{KL}(q(\mathbf{z}|\mathbf{x})||p(\mathbf{z}|y)) \right ] \\
	  & = \nabla_{\phi} \mathbb{E}_{Gumbel(g|0,1)} \left [ D_{KL}(q(\mathbf{z}|\mathbf{x})||p(\mathbf{z}|f(\phi, g, \tau)) \right ]  \\
	  & = \mathbb{E}_{Gumbel(g|0,1)} \left [ \nabla_{\phi} D_{KL}(q(\mathbf{z}|\mathbf{x})||p(\mathbf{z}|f(\phi, g, \tau)) \right ] , \\
    \end{aligned}
\end{equation}

where $f(\phi, g, \tau)$ is defined as:

\begin{equation} \label{gumbel_conditional_normalization}
\setlength\abovedisplayskip{0pt}
\centering
    \displaystyle
    \begin{aligned} 
    	  f_{k}(\phi,g,\tau)_{1 \leq k \leq K} & = \frac{\textrm{exp} \, (( \boldsymbol{\pi}_{\phi}(\mathbf{z})_{k} + g_{k} )/\tau))}{\sum_{i=1}^{K} \textrm{exp} \, (( \boldsymbol{\pi}_{\phi}(\mathbf{z})_{i} + g_{i} )/\tau)}  \\
    	  & = \frac{\textrm{exp} \, (( \alpha_{k} + g_{k} )/\tau))}{\sum_{i=1}^{K} \textrm{exp} \, (( \alpha_{i} + g_{i} )/\tau)}
    \end{aligned}
\end{equation}

By means of the Gumbel-Max \cite{jang2016categorical, maddison2016concrete}, we are able to reparameterize the categorical distribution $q_{\phi}(y|\mathbf{z})$ as a deterministic function $f$ of $\phi$ and a random variable $g \sim Gumbel(0,1)$, where $g$ does not depend on $\phi$. However, Gumbel-Max definition comprehend an $arg\_max$ operation, which derivative is $0$ everywhere except at the boundary of state changes, where it is not defined. Therefore we use the Gumbel-Softmax relaxation \cite{jang2016categorical, maddison2016concrete} as defined in $f(\phi,g,\tau)$ in order to take gradients of $D_{KL}$ with respect to $\phi$. Due to the relaxation, $p(\mathbf{z}|f(\phi, g, \tau)$ is a mixture of Gaussians. In order to get a Kullback-Leibler divergence between two gaussian distributions, which have an analytical form as shown in equation \ref{kl_gaussians}, we use the Straight-Through Gumbel estimator as defined in \cite{jang2016categorical}, which discretize $y$ using $arg\_max$ during forward pass, while using the continuous approximation in the backward pass. The final expectation over gradients follows: 

\begin{equation} \label{gumbel_conditional_max_}
\setlength\abovedisplayskip{0pt}
\centering
    \displaystyle
    \begin{aligned} 
    	  \mathbb{E}_{Gumbel(g|0,1)} & \left [ \nabla_{\phi}  D_{KL}(\mathcal{N}(\boldsymbol{\mu}_{\phi}, \boldsymbol{\sigma}^{2}_{\phi}\mathbf{I})||\mathcal{N}(\boldsymbol{\mu}_{y}, \boldsymbol{\sigma}^{2}_{y}\mathbf{I}) \right ] , \\
    	  & y = \textrm{arg\_max}(f(\phi, g, \tau)) \\
    \end{aligned} 
\end{equation}

While this makes our estimator a biased one, we see it works well in practice. Following the same procedure, we can derive an expectation over gradients for the second term in equation \ref{loss_unsup_kl}:

\begin{equation} \label{gumbel_prior_reparametrization_}
\setlength\abovedisplayskip{0pt}
\centering
    \displaystyle
    \begin{aligned} 
      & \nabla_{\phi} \mathbb{E}_{q_{\phi}(\mathbf{z}|\mathbf{x})} \left [ D_{KL}(q(y|\mathbf{z})||p(y))\right ] \\
      & = \nabla_{\phi} \mathbb{E}_{\mathcal{N}(\epsilon|0,1)} \left [ D_{KL}(\boldsymbol{\pi}_{\phi}(f(\mathbf{x},\boldsymbol{\epsilon}))||p(y)) \right ] \\
      & = \mathbb{E}_{\mathcal{N}(\epsilon|0,1)} \left [ \nabla_{\phi} D_{KL}(\boldsymbol{\pi}_{\phi}(f(\mathbf{x},\boldsymbol{\epsilon}))||p(y)) \right ] , \\
      \end{aligned}
\end{equation}

where:

\begin{equation} \label{gaussian_reparametrization}
\setlength\abovedisplayskip{0pt}
\centering
    \displaystyle
    \begin{aligned} 
      f(\mathbf{x},\boldsymbol{\epsilon}) = \boldsymbol{\mu}_{\phi}(\mathbf{x}) + \boldsymbol{\sigma}^{2}_{\phi}(\mathbf{x}) \odot \boldsymbol{\epsilon}
      \end{aligned}
\end{equation}

{\small
\bibliographystyle{ieee}
\bibliography{supplementary_arxiv}
}

\end{document}